\documentclass{article}

\usepackage[final]{nips_2018}
\usepackage{amsmath}

\usepackage[utf8]{inputenc} 
\usepackage[T1]{fontenc}    
\usepackage{hyperref}       
\usepackage{url}            
\usepackage{booktabs}       
\usepackage{amsfonts}       
\usepackage{nicefrac}       
\usepackage{microtype}      

\usepackage{algorithm}
\usepackage[noend]{algpseudocode}
\usepackage{graphicx}   
\usepackage{multirow}
\usepackage{mathtools,xparse}

\DeclarePairedDelimiter{\norm}{\lVert}{\rVert}
\NewDocumentCommand{\normL}{ s O{} m }{%
  \IfBooleanTF{#1}{\norm*{#3}}{\norm[#2]{#3}}%
}
\DeclareMathOperator*{\argmin}{arg\,min}
\newcommand{\Tau}{\mathrm{T}}

\title{Gradient Agreement as an Optimization Objective for Meta-Learning}

\author{Amir Erfan Eshratifar,\textsuperscript{1}
David Eigen,\textsuperscript{2}
Massoud Pedram,\textsuperscript{1}\\
\textsuperscript{1}{Department of Electrical Engineering at University of Southern California, Los Angeles, CA 90089, USA}\\
\textsuperscript{2}{Clarifai, San Francisco, CA 94105, USA}\\
eshratif@usc.edu,
deigen@clarifai.com,
pedram@usc.edu}
\begin{document}

\maketitle

\begin{abstract}
This paper presents a novel optimization method for maximizing generalization over tasks in meta-learning. The goal of meta-learning is to learn a model for an agent adapting rapidly when presented with previously unseen tasks. Tasks are sampled from a specific distribution which is assumed to be similar for both seen and unseen tasks. We focus on a family of meta-learning methods learning initial parameters of a base model which can be fine-tuned quickly on a new task, by few gradient steps (MAML). Our approach is based on pushing the parameters of the model to a direction in which tasks have more agreement upon. If the gradients of a task agree with the parameters update vector, then their inner product will be a large positive value. As a result, given a batch of tasks to be optimized for, we associate a positive (negative) weight to the loss function of a task, if the inner product between its gradients and the average of the gradients of all tasks in the batch is a positive (negative) value. Therefore, the degree of the contribution of a task to the parameter updates is controlled by introducing a set of weights on the loss function of the tasks. Our method can be easily integrated with the current meta-learning algorithms for neural networks. Our experiments demonstrate that it yields models with better generalization compared to MAML and Reptile.
\end{abstract}

\section{Introduction}

Many successes of deep learning today rely on enormous amounts of labeled data, which is not practical for problems with small data, in which acquiring enough labeled data is expensive and laborious. Moreover, in many mission critical applications, such autonomous vehicles and drones, an agent needs to adapt rapidly to unseen environments. The ability to learn rapidly is a key characteristic of humans that distinguishes them from artificial agents. Humans leverage prior knowledge learned earlier when adapting to a changing task, an ability that can be explained as Bayesian inference~(\cite{humanlevel}). Pioneered by (\cite{Schmidhuber}), meta-learning has emerged recently as a novel field of study for learning from small amounts of data. Meta-learning algorithms learn their base models by sampling many different smaller tasks from a large data source instead trying to emulate a computationally intractable Bayesian inference. As a result, one might expect that the meta-learned model is capable of generalizing well to new unseen tasks because of task-agnostic way of training. 

In this paper, we focus on a family of meta-learning algorithms that aim to learn the initialization of a network, which is then fine-tuned at test time on the new task by few steps of gradient descent. The most prominent work in this family is MAML (\cite{MAML}). MAML directly optimizes the performance of the model with respect to the test time fine-tuning procedure leading to a better initialization point. 

In the parallel or batched version of MAML style algorithms (e.g. Reptile~(\cite{Reptile}), at each iteration we optimize on $N$ tasks. However, the contribution of each task to the model parameter updates is assumed to be the same, i.e., the update rule $\theta \leftarrow \theta - \beta \nabla \sum L_{\tau_i}(f_{\theta_i})$ is merely the average of the tasks losses. As an example, suppose a scenario in which a batch of $N$ tasks is sampled and $N-1$ tasks have strong agreement on the gradient direction but one task has a large disagreeing gradient. In this case, the disagreeing task can adversely affect the next model parameter update.

The contribution of this paper is introducing an optimization objective, i.e., gradient agreement by which a model can generalize better across tasks. The proposed algorithm and its mathematical analysis are illustrated in Sections~\ref{gradient_agreement_section} and~\ref{analysis_sec}, respectively. In Section~\ref{results_section}, we provide some empirical insights by evaluations on miniImageNet~(\cite{MatchingNetwork}) and Omniglot~(\cite{omniglot}) classification tasks, and a toy example regression task.

\section{Gradient Agreement} \label{gradient_agreement_section}
In this work, we consider the optimization problem of MAML which is finding a good initial set of parameters, $\theta$. The goal is that for a randomly sampled task $\tau$ with corresponding loss $L_\tau$, the model converges to a low loss point after $k$ number of gradient steps. This objective can be presented as $min_\theta \mathop{\mathbb{E}}[L_\tau (U_\tau^k(\theta))]$, where $U_\tau^k(\theta)$ is a function performing $k$ gradient steps that updates $\theta$ by samples drawn from task $\tau$. Minimizing this objective function, requires two steps in MAML style works: 1) Adapting
to a new task $T_i$ in which the model parameters $\theta$ become $\theta_i$ by $k$ steps of gradient descent on task $T_i$, referred to as inner-loop optimization, 2) Updating the model parameters by optimizing for the performance
of $f_{\theta_i}$ with respect to $\theta$ across all sampled tasks performed as follows: $\theta \leftarrow \theta - \beta \nabla \sum_i L_{\tau_i}(f_{\theta_i})$, and referred to as outer-loop optimization.

\begin{algorithm}
\caption{Gradient Agreement Algorithm}\label{gradient_agreement_algorithm}
\begin{algorithmic}[1]  
	\State {Initialize model parameters, $\theta$}
     \For{\texttt{iteration = 1,2,...}}
        \State Sample a batch of $N$ tasks $\tau_i$ $\sim$ p($\tau$)
        \For{all $\tau_i$}
        \State Split the examples of the task into $k$ sub-batches
        \State $\theta_i$ = $\theta$ - $\alpha_{inner} \nabla L_{\tau_i}(f_\theta)$ for $k$ steps
        \State $g_i$ = $\theta$ - $\theta_i$
        \EndFor
        \State $w_i = \frac{\sum_{j \in \Tau } (g_i^T g_j) }{\sum_{k \in \Tau} |\sum_{j \in \Tau } (g_k^T g_j)|} $ for all i
        \State $\theta_{new}$ = $\theta$ - $\alpha_{outer}$ $\sum_{i} w_i \nabla L_{\tau_i}(f_{\theta_i})$ (MAML)
        \State $\theta_{new}$ = $\theta$ + $\alpha_{outer}$ $\sum_{i}$ $w_i$ ($\theta_i$ - $\theta$) (Reptile)
      \EndFor
	\State \Return $\theta$
\end{algorithmic}
\end{algorithm}

The assumption in the prior arts~(\cite{MAML,Reptile}) is that each task should contribute equally for the direction of the outer-loop optimization step. However, in our approach, we associate a weight to the loss of each task, and the meta-optimization rule will become: $\theta \leftarrow \theta - \beta \nabla \sum w_i L_{\tau_i}(f_{\theta_i})$. Assuming that the gradient update vector for each task is presented by $g_i$, we define $w_i$ as:

\begin{equation}\label{weights_formula}
w_i = \frac{\sum_{j \in \Tau } (g_i^T g_j) }{\sum_{k \in \Tau} |\sum_{j \in \Tau } (g_k^T g_j)|} \sim g_i^T g_{avg}
\end{equation}
The proposed $w_i$ is proportional to the inner product of the gradient of a task and the average of the gradients of all tasks in a batch. Therefore, if the gradient of a task is aligned with the average of the gradients of all tasks, it contributes more than other tasks to the next parameter update. With this insight, the proposed optimization method, pushes the model parameters towards initial parameters that more tasks agree upon. The full procedure of the proposed optimization is illustrated in Algorithm~\ref{gradient_agreement_algorithm}.

\section{Analysis} \label{analysis_sec}
In this section, inspired by~(\cite{Bilevel}), we provide a mathematical analysis for the proposed method. Denoting a batch of sampled tasks as $\Tau$, the outer-loop update in batched version of MAML with current parameters $\theta^t$ will be:
\begin{equation}
\theta^{t+1} = \theta^t - \beta \sum_{i \in \Tau } \nabla L_{\tau_i}(\theta_i) 
\end{equation}
Instead, in gradient agreement approach, we look for a linear combination of task losses that lead to a better approximation of the tasks errors. We use $w_i$ per each task which is estimated at each iteration. Therefore, the goal is to find parameters $\theta$ and coefficients $w_i$ so that the model performs well on the sampled batch of tasks. We thus aim to optimize the following objective function:

\begin{equation}\label{ga_objective}
\hat{\theta}, \hat{\textbf{w}} = \argmin_{\theta, \textbf{w}} \sum_{i \in \Tau } L_{\tau_i}(\theta(\textbf{w})) + \frac{\mu}{2} \norm{\textbf{w}}_2^2
\end{equation}

\begin{equation}\label{constraints}
subj.~to:~~\theta(\textbf{w}) = \argmin_{\Bar{\theta}} \sum_{i \in \Tau } w_i L_{\tau_i}(\Bar{\theta}), ~~ \norm{\textbf{w}}_1 = 1
\end{equation}

where, $\textbf{w}$ is the vector of all weights, $w_i$ and $i \in \Tau$, associated with each task, and $\mu$ is a regularization parameter for the distribution of weights. In this section we show that the introduced inner product metric in \ref{weights_formula} is corresponding to minimizing the objective function of Equation~\ref{ga_objective}. Note that the solution of Equation~\ref{ga_objective} does not change if we multiply all the weights, $w_i$, by the same positive constant. As a result, we normalize the magnitude of $\textbf{w}$ to one. A typical method to solve the above optimization is the so-called \textit{implicit differentiation}~(\cite{implicit_differentiation}) method to solve a linear system in the second order derivatives of loss functions, which leads to solving a very high-dimensional linear system. To avoid computational complexities of the aforementioned method, we apply proximal approximation using Taylor series.

The loss function of $i$-th task at $\theta^t$ can be approximated using Taylor series as:
\begin{equation}\label{first_order_approximation}
L_{\tau_i}(\theta) \approx L_{\tau_i}(\theta^t) + g_i (\theta - \theta^t)
\end{equation}
where $g_i$ is the update vector obtained in the inner-loop optimization for $i$-th task. By plugging in the first order approximation of~(\ref{first_order_approximation}) to~(\ref{ga_objective}) and (\ref{constraints}), we obtain the following objective function:

\begin{equation}\label{ga_objective_modified}
\theta^{t+1}, \hat{\textbf{w}} = \argmin_{\theta, \textbf{w}} \sum_{i \in \Tau } [L_{\tau_i}(\theta^t) + g_i (\theta(\textbf{w}) - \theta^t)] + \frac{\mu}{2} \norm{\textbf{w}}_2^2
\end{equation}

\begin{equation}\label{lower_level_modified}
subj.~to:~~\theta(\textbf{w}) = \argmin_{\Bar{\theta}} \sum_{i \in \Tau } w_i [L_{\tau_i}(\theta^t) + g_i (\Bar{\theta} - \theta^t)] + \frac{\norm{\Bar{\theta} - \theta^t}_2^2}{2\epsilon}, ~~ \norm{\textbf{w}}_1 = 1
\end{equation}

The closed-form solution to the quadratic problem in~(\ref{lower_level_modified}) is a classical SGD update rule of:
\begin{equation}\label{lower_level_modified_solution}
\theta(\textbf{w}) = \theta^t - \epsilon \sum_{i \in \Tau } w_i g_i 
\end{equation}
Now, we plug in the solution in~(\ref{lower_level_modified_solution}) to (\ref{ga_objective_modified}):

\begin{equation}\label{ga_objective_modified2}
\hat{\textbf{w}} = \argmin_{\textbf{w}} \sum_{i \in \Tau } \sum_{j \in \Tau } (-\epsilon  w_i g_i^T g_j) + \frac{\mu}{2} \norm{\textbf{w}}_2^2
\end{equation}

We compute the derivative of (\ref{ga_objective_modified2}) with respect to $w_i$ for all $i \in T$ and set it to zero, temporarily ignoring the normalization constraint on $\textbf{w}$:

\begin{equation}\label{derivative_equal-zero}
\sum_{j \in \Tau } (-\epsilon g_i^T g_j) + \mu w_i = 0
\end{equation}

Therefore, $w_i$ will be: 
\begin{equation}\label{w_found}
w_i = \lambda \sum_{j \in \Tau } (g_i^T g_j) 
\end{equation}

Now, by applying the $L_1$ constraint on the norm of $\textbf{w}$, we obtain $\lambda$ as:

\begin{equation}\label{w_found}
\sum_{i \in \Tau }|w_i| = 1 \Rightarrow \lambda \sum_{i \in \Tau} | \sum_{j \in \Tau } (g_i^T g_j)| = 1 \Rightarrow \lambda = \frac{1}{\sum_{i \in \Tau} |\sum_{j \in \Tau } (g_i^T g_j)|}
\end{equation}

As a result, $w_i$ will be:

\begin{equation}\label{w_found}
w_i = \frac{\sum_{j \in \Tau } (g_i^T g_j) }{\sum_{k \in \Tau} |\sum_{j \in \Tau } (g_k^T g_j)|} 
\end{equation}

\section{Experiments and Results} \label{results_section}

\subsection{One-Dimensional Sine Wave Regression}

As a simple case study, consider the 1D sine wave regression problem. A task $\tau = (a, \phi)$ is defined by the amplitude $a$ and phase $\phi$ of a sine wave function
$f_{\tau} (x) = a sin(x + b)$, $a$ and $\phi$ are sampled from $U([0.1, 5.0])$ and $U([0, 2\pi])$, respectively. The average of $L_2$ loss over 1000 unseen sine wave tasks is 0.13 and 0.08 for Reptile and gradient agreement method, respectively.

\begin{figure}[!htb]
\minipage{0.33\textwidth}
  \includegraphics[width=\linewidth]{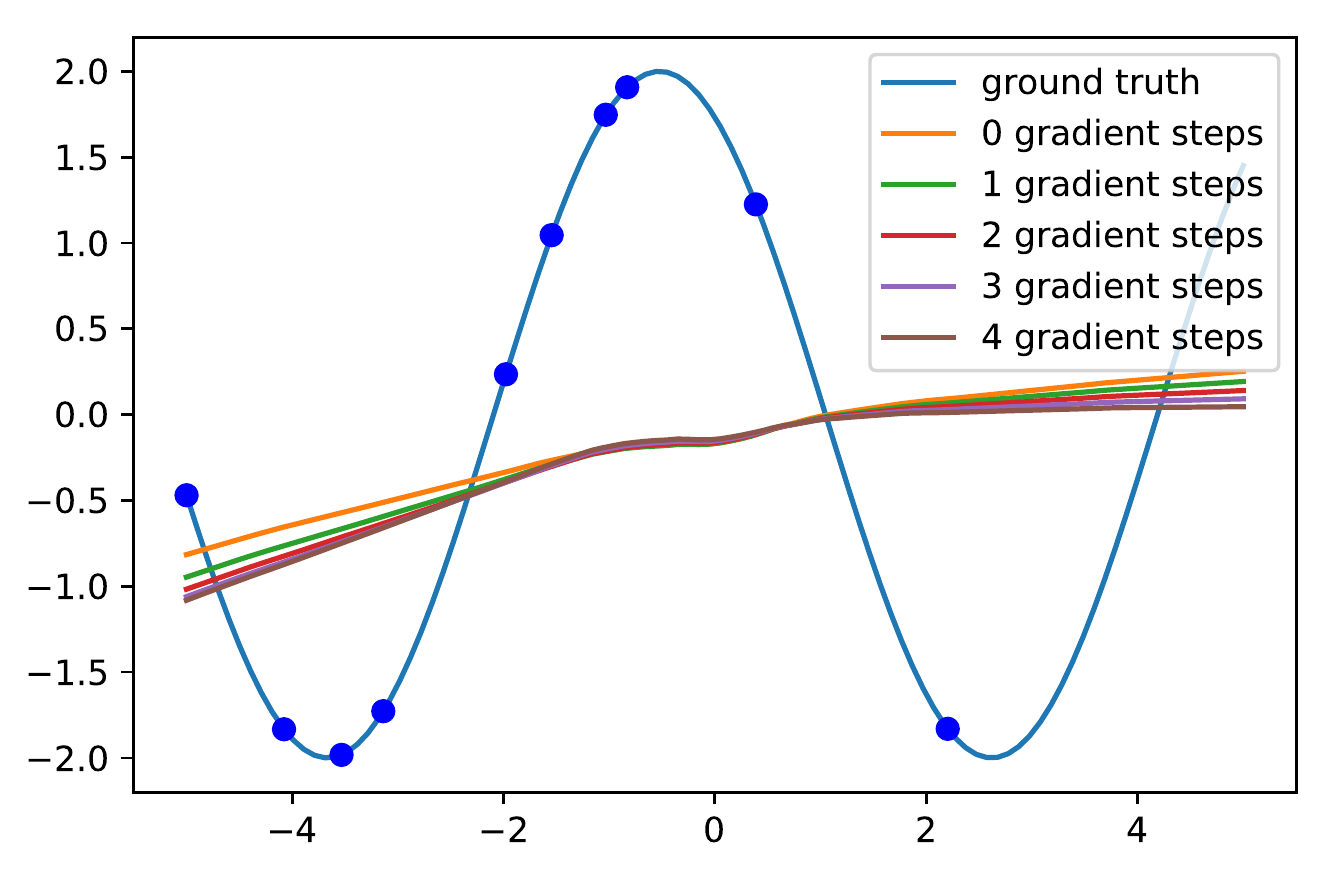}
\endminipage\hfill
\minipage{0.33\textwidth}
  \includegraphics[width=\linewidth]{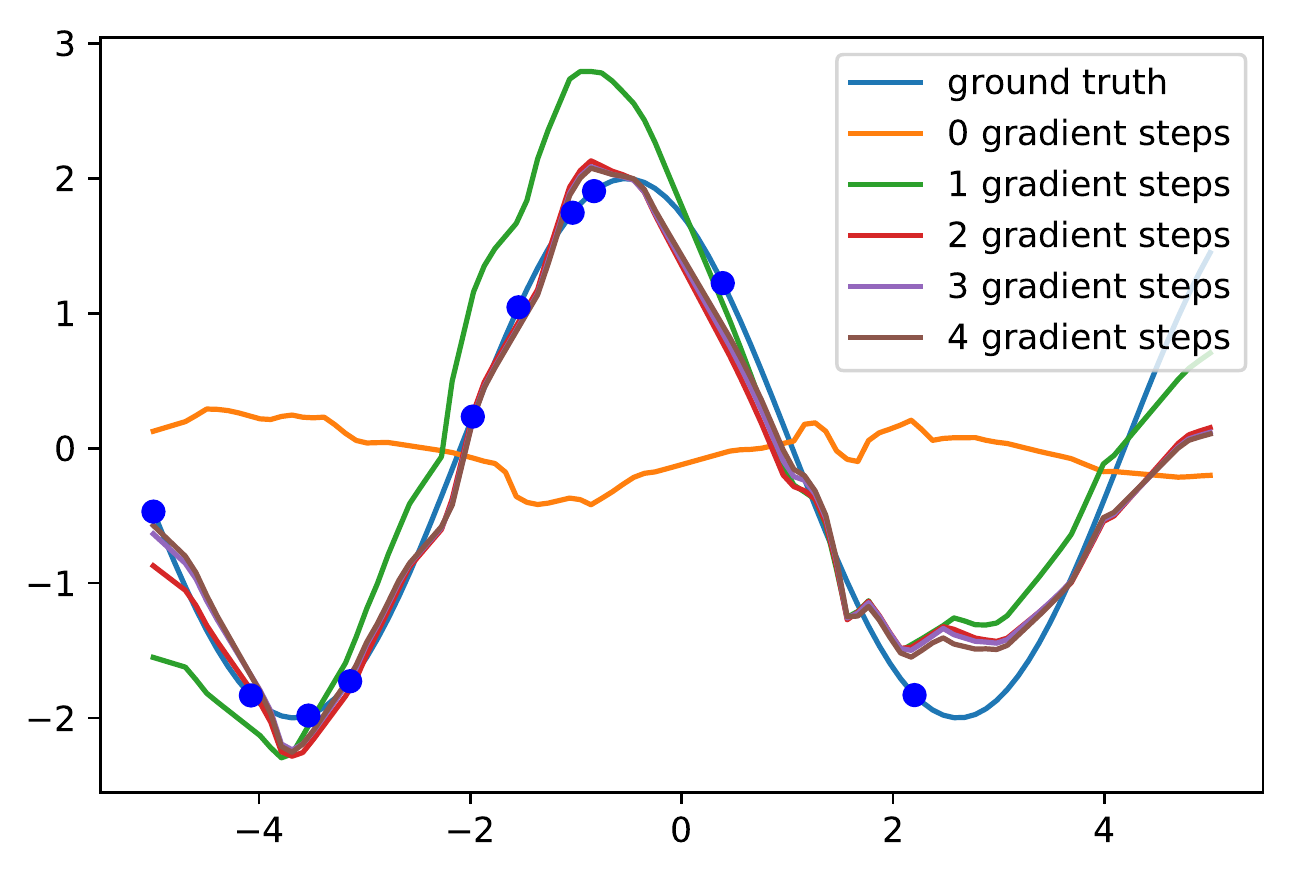}
\endminipage\hfill
\minipage{0.33\textwidth}%
  \includegraphics[width=\linewidth]{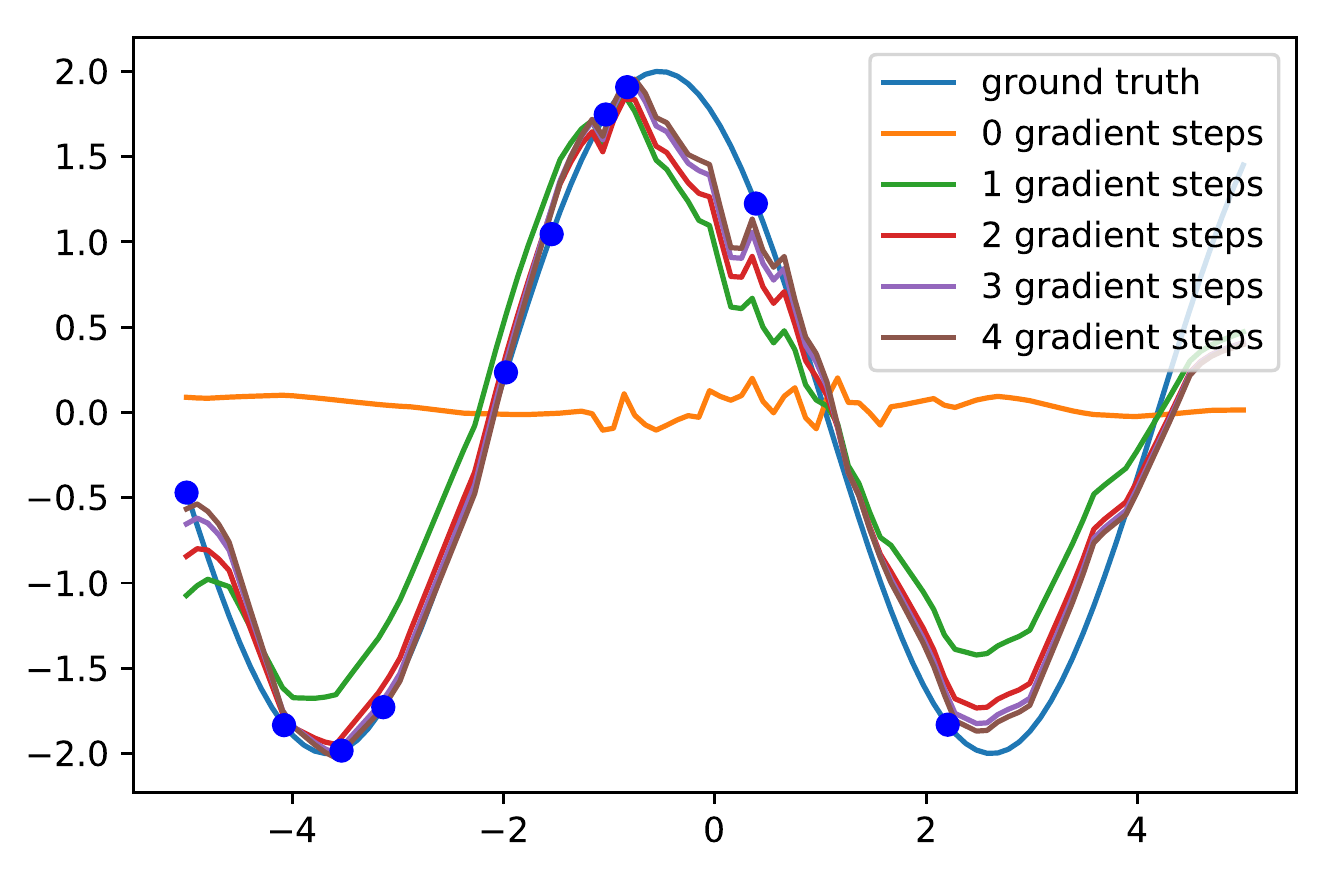}
\endminipage
\caption{Demonstration of Reptile and our method on a toy few-shot regression problem, where we train on 10 sampled points of a sine wave, performing 5 gradient steps on a MLP with layers 1 $\rightarrow$ 64 $\rightarrow$ 64 $\rightarrow$ 1. left: before training, middle: after Reptile training, right: after gradient agreement based training.}
\end{figure}

\subsection{Few-shot Classification}

We evaluate our method on two popular few-shot classification tasks, miniImageNet~(\cite{MatchingNetwork}) and Omniglot~(\cite{omniglot}). The Omniglot dataset consists of 20 instances of 1623 characters from 50 different alphabets. The miniImagenet dataset includes 64 training classes, 12 validation classes, and 24 test classes, each class contains 600 examples. The problem of N-way classification
is set up as follows: we sample N classes from the total C classes in the dataset and then selecting K examples for each class.
For our experiments, we use the same CNN architectures and data pre-processing as (\cite{MAML}) and (\cite{Reptile}). The results for miniImageNet and Omniglot are presented in Tables~\ref{tab:miniimagenet_results} and ~\ref{tab:omniglot_results}, respectively. The models learned by our approach compares well to the state-of-the-art results on these two tasks. It substantially outperforms the prior methods on the challenging task of miniImageNet.

\begin{table}[h]
	\caption{Accuracy results on miniImageNet.}
	\centering 
	\begin{tabular}{|c|c|c|} \hline
	\textbf{Algorithm} & \textbf{5way 1shot} & \textbf{5way 5shot} \\ 
	\hline
	MAML~(\cite{MAML})	& 	48.70$\pm$1.84\% 	&	63.11$\pm$0.92\% \\
	\hline
	Reptile~(\cite{Reptile})	&   49.97$\pm$0.32\%		&	65.99$\pm$0.58\% \\ \hline
    Gradient Agreement	& 54.80$\pm$1.19\%	& 73.27$\pm$0.62\%\
    \\ 
    \hline
\end{tabular}
\label{tab:miniimagenet_results} 
\end{table}

\begin{table}[h]
	\caption{Accuracy results on Omniglot.}
	\centering 
	\begin{tabular}{|c|c|c|c|c|} \hline
	\textbf{Algorithm} & \textbf{5way 1shot} & \textbf{5way 5shot} & \textbf{20way 1shot} & \textbf{20way 5shot} \\ 
	\hline
	MAML~(\cite{MAML})	& 	98.7$\pm$0.4\% 	&	99.9$\pm$0.1\% &	95.8$\pm$0.3\% &	98.9$\pm$0.2\%\\
	\hline
	Reptile~(\cite{Reptile})	&   97.68$\pm$0.04\%		&	99.48$\pm$0.06\% &	89.43$\pm$0.14\% &	97.12$\pm$0.32\% \\ 
	\hline
    Gradient Agreement	& 98.6$\pm$0.63\%	& 99.8$\pm$0.12\% &	94.25$\pm$0.51\% &	98.99$\pm$1.34\%
    \\ 
    \hline
\end{tabular}
\label{tab:omniglot_results} 
\end{table}

\section{Conclusion and Future Works}
In this work, we presented a generalization method for meta-learning algorithms that adjusts the model parameters by introducing a set of weights over the loss functions of tasks in a batch in order to maximize the dot products between the gradients of different tasks. The higher the inner products between the gradients of different tasks, the more agreement they have upon the model parameters update. We also presented the objective function of this optimization method by a theoretical analysis using first order Taylor series approximation. This geometrical interpretation of optimization in meta-learning studies can be an interesting future direction.
\medskip

\small
\bibliographystyle{unsrtnat}
\bibliography{references.bib}

\begin{thebibliography}{8}
\providecommand{\natexlab}[1]{#1}
\providecommand{\url}[1]{\texttt{#1}}
\expandafter\ifx\csname urlstyle\endcsname\relax
  \providecommand{\doi}[1]{doi: #1}\else
  \providecommand{\doi}{doi: \begingroup \urlstyle{rm}\Url}\fi

\bibitem[Brenden M~Lake and Tenenbaum(2015)]{humanlevel}
Ruslan~Salakhutdinov Brenden M~Lake and Joshua~B Tenenbaum.
\newblock Human-level concept learning through probabilistic program induction.
\newblock In \emph{Science}, 2015.

\bibitem[Schmidhuber(1987)]{Schmidhuber}
Jurgen Schmidhuber.
\newblock Evolutionary principles in self-referential learning.
\newblock Master's thesis, 1987.

\bibitem[Finn et~al.(2017)Finn, Abbeel, and Levine]{MAML}
Chelsea Finn, Pieter Abbeel, and Sergey Levine.
\newblock Model-agnostic meta-learning for fast adaptation of deep networks.
\newblock \emph{CoRR}, abs/1703.03400, 2017.
\newblock URL \url{http://arxiv.org/abs/1703.03400}.

\bibitem[Nichol et~al.(2018)Nichol, Achiam, and Schulman]{Reptile}
Alex Nichol, Joshua Achiam, and John Schulman.
\newblock On first-order meta-learning algorithms.
\newblock \emph{CoRR}, abs/1803.02999, 2018.
\newblock URL \url{http://arxiv.org/abs/1803.02999}.

\bibitem[Vinyals et~al.(2016)Vinyals, Blundell, Lillicrap, Kavukcuoglu, and
  Wierstra]{MatchingNetwork}
Oriol Vinyals, Charles Blundell, Timothy~P. Lillicrap, Koray Kavukcuoglu, and
  Daan Wierstra.
\newblock Matching networks for one shot learning.
\newblock \emph{CoRR}, abs/1606.04080, 2016.
\newblock URL \url{http://arxiv.org/abs/1606.04080}.

\bibitem[Lake et~al.(2011)Lake, Salakhutdinov, Gross, and Tenenbaum]{omniglot}
Brenden~M. Lake, Ruslan Salakhutdinov, Jason Gross, and Joshua~B. Tenenbaum.
\newblock One shot learning of simple visual concepts.
\newblock In \emph{CogSci}, 2011.

\bibitem[Jenni and Favaro(2018)]{Bilevel}
Simon Jenni and Paolo Favaro.
\newblock Deep bilevel learning.
\newblock volume abs/1809.01465, 2018.

\bibitem[Domke(2012)]{implicit_differentiation}
Justin Domke.
\newblock Generic methods for optimization-based modeling.
\newblock In \emph{Proceedings of the Fifteenth International Conference on
  Artificial Intelligence and Statistics}, 2012.

\end{thebibliography}

\end{document}